\pdfoutput=1

\documentclass[11pt]{article}

\usepackage[]{acl}

\usepackage{times}
\usepackage{latexsym}
\usepackage{amsmath}
\usepackage{graphics}
\usepackage[T1]{fontenc}

\usepackage[utf8]{inputenc}

\usepackage{microtype}
\usepackage{graphicx}
\usepackage{multirow}
\usepackage{threeparttable}

\makeatletter
\newcommand{\rnum}[1]{\expandafter\@slowromancap\romannumeral #1@}
\makeatother

\usepackage{booktabs}

\title{Probing Simile Knowledge from Pre-trained Language Models}

\author{
Weijie Chen\textsuperscript{1}\thanks{ \ \ Equal contribution. Work is done during Weijie's internship at NetEase Inc..},
Yongzhu Chang\textsuperscript{2*},
Rongsheng Zhang\textsuperscript{2*},
Jiashu Pu\textsuperscript{2} \\
\bf{Guandan Chen\textsuperscript{2}, 
Le Zhang\textsuperscript{2}, 
Yadong Xi\textsuperscript{2},
Yijiang Chen\textsuperscript{1} and 
Chang Su\textsuperscript{1}}\thanks{ \ \ Corresponding author. E-mail: suchang@xmu.edu.cn.}
\\
\textsuperscript{1} School of Informatics, Xiamen University, Xiamen, China 
\\
\textsuperscript{2} Fuxi AI Lab, NetEase Inc., Hangzhou, China
\\
\small \texttt{chenwj@stu.xmu.edu.cn},
\texttt{\{changyongzhu,zhangrongsheng\}@corp.netease.com}
}

\begin{document}
\maketitle

\begin{abstract}
Simile interpretation (SI) and simile generation (SG) are challenging tasks for NLP because models require adequate world knowledge to produce predictions. Previous works have employed many hand-crafted resources to bring knowledge-related into models, which is time-consuming and labor-intensive. In recent years, pre-trained language models (PLMs) based approaches have become the de-facto standard in NLP since they learn generic knowledge from a large corpus. The knowledge embedded in PLMs may be useful for SI and SG tasks. Nevertheless, there are few works to explore it. In this paper, we probe simile knowledge from PLMs to solve the SI and SG tasks in the unified framework of simile triple completion for the first time. The backbone of our framework is to construct masked sentences with manual patterns and then predict the candidate words in the masked position. In this framework, we adopt a secondary training process (Adjective-Noun mask Training) with the masked language model (MLM) loss to enhance the prediction diversity of candidate words in the masked position. Moreover, pattern ensemble (PE) and pattern search (PS) are applied to improve the quality of predicted words. Finally, automatic and human evaluations demonstrate the effectiveness of our framework in both SI and SG tasks.
\end{abstract}

\section{Introduction}
\label{sec:introduction}

The simile, which is a special type of metaphor, is defined as a figurative expression in which two fundamentally different things are explicitly compared, usually using ``like'' or ``as'' \cite{israel2004simile,simile_sjs}. 
It is widely used in literature because it can inspire the reader's imagination \cite{pual} by giving a vivid and unexpected analogy between two objects with similar attributes.
A simile sentence usually contains three key elements: the \textbf{tenor}, the \textbf{attribute} and the \textbf{vehicle},\footnote{Tenor: the logical subject of the comparison, usually a noun phrase. Attribute: what things being compared have in common, usually an adjective. Vehicle: the logical object of the comparison, usually a noun phrase.} which can be defined in the form of a triple (\textit{tenor}, \textit{attribute}, \textit{vehicle}) \cite{SongGFLL21}.
For example, the simile sentence ``Love is as thorny as rose'' can be extracted as the triple (love, thorny, rose), where the tenor is ``love'', the vehicle is ``rose'', and the attribute is ``thorny''. 
Note that a simile triple can produce different simile sentences with different templates. For the example triple above, the simile sentences can be also constructed as ``\textit{love is thorny like rose}" with the pattern ``\textit{tenor} is \textit{attribute} like \textit{vehicle}".

\begin{figure}[t]
\centering
\includegraphics[width=\linewidth]{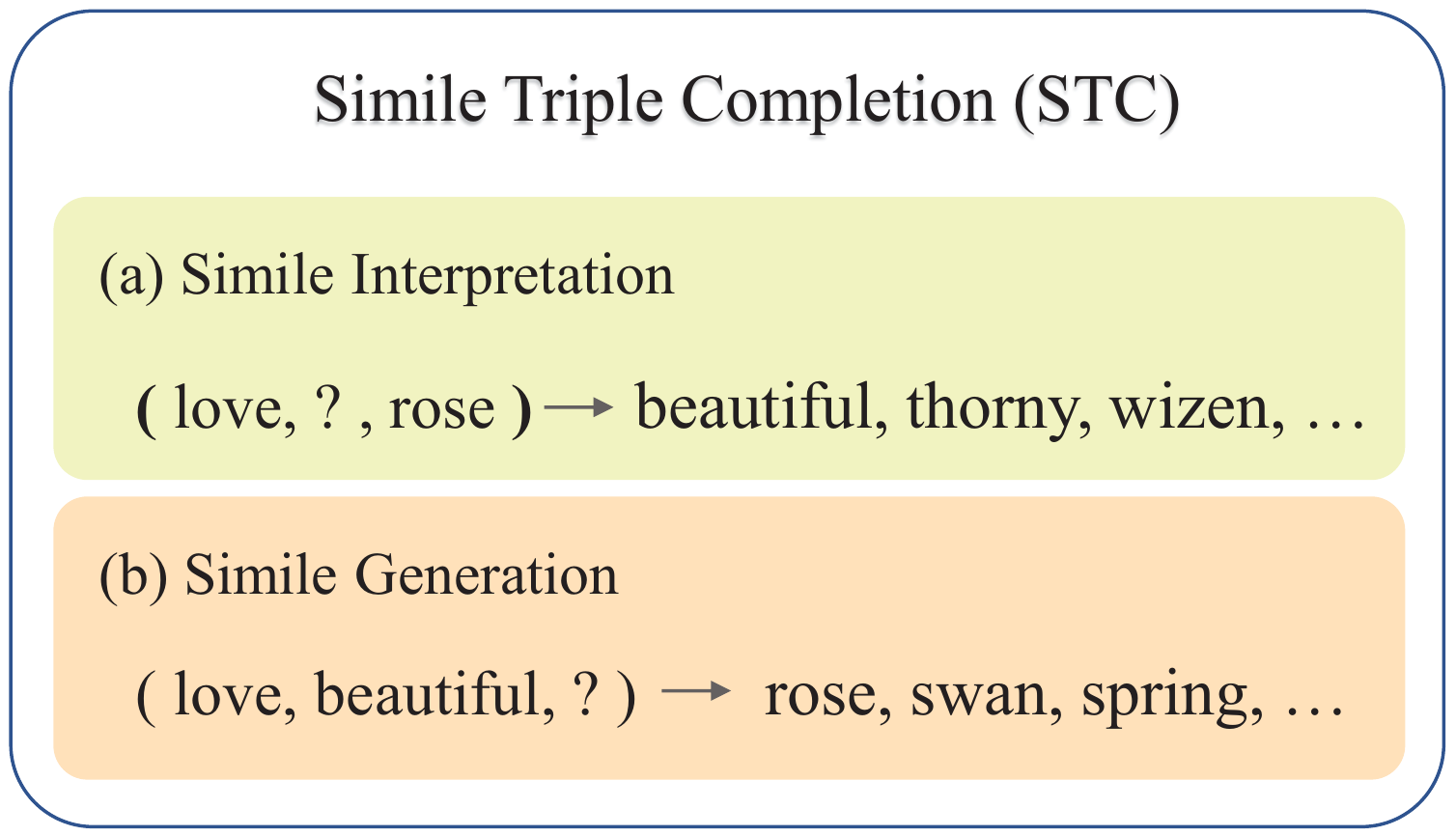}
\caption{In the form of triple, the tasks of Simile Interpretation and Simile Generation can be unified into Simile Triple Completion.}
\label{fig1}
\end{figure}

The study of simile is benefit to many downstream tasks, like sentiment analysis \cite{RentoumiVKM12}, question answering \cite{xiaobing}, writing polishment \cite{xiaomi} and creative writing \cite{Metaphoria}. \textbf{Simile interpretation} (SI) \cite{QadirRW16, SuTC16} and \textbf{simile generation} (SG) \cite{Yu2019HowTA} are the two important tasks in the study of simile \cite{TongSL21}.
The SI task is to find suitable attributes as a mediator between the tenor and vehicle. Likewise, the SG task is to select a proper vehicle for the tenor with the given attribution. And these two tasks can be unified into the form of \textbf{simile triple completion} (STC) \citep{SongGFLL21} as shown in Figure \ref{fig1}.


Previous works on the SI and SG tasks relied on a limited training corpus or labor-intensive knowledge base, which leads to an upper limit on the diversity of results.
\citep{SongGFLL21} collected sentences containing comparator words from a Chinese essays corpus and manually annotated them to obtain the simile triple. Some works \citep{StoweCPMG20, Metaphoria,2016VealeSK} relied on a knowledge base such as ConceptNet\footnote{https://conceptnet.io/}, FrameNet\footnote{https://framenet.icsi.berkeley.edu/fndrupal/}, which are scarce to other languages because it is time-consuming and labor-intensive to construct such a knowledge base. 
It is notable that pre-trained language models (PLMs) \citep{devlin2018bert,gpt2} have made significant progress recently in many NLP tasks since it learns generic knowledge such as grammar, common sense from a large corpus \citep{DavisonFR19, prompt_survey, gpttoo}. Considering the sufficient existence of simile in the large corpus, it's reasonable to assume that PLMs are equipped with rich knowledge of similes during the pre-training stage. 
However, few works have explored directly probing the knowledge of simile from the PLMs.

In this paper, we propose a unified framework to solve the SI and SG tasks by mining the knowledge in PLMs, which does not require fine-labeled training data or knowledge graphs. The backbone of our method is to construct masked sentences with manual patterns from an incomplete simile triple, and then use language models with MLM heads to predict the masked words over the task-specific vocabulary. We take the $K$ words with the highest probability as the result words. However, there are problems with this crude approach.
Firstly, the predicted words should be creative and surprised for the simile sentence. On the contrary, the PLMs tend to predict common words (e.g., good, bad) with a higher probability. To address this issue, we introduce a secondary pre-training stage - Adjective-Noun mask Training (ANT), where only the noun or adjective contained in the \textit{amod} dependencies \citep{Uni_Depend} could be masked in the MLM training process and the number of words masked times are limited.  Secondly, the words predicted by MLM have a preference for different patterns.  For this reason, we employ a pattern ensemble to obtain high-quality and robust results.  Finally, we also introduce a prompt-search method to improve the quality of the simile component predictions.

Our main contributions are as follows: 
\begin{itemize}
\item We propose a unified framework to solve both the simile interpretation (SI) and simile generation (SG) tasks based on pre-trained models. To the best of our knowledge, it is the first work to introduce pre-trained language models to unify these tasks. 
\item We propose a secondary pre-training stage that effectively improves the prediction diversity. Further, we employ the pattern-ensemble and pattern-search approaches to obtain better results.
\item We compare our models on both automated metrics and manual measures, and the results show that our approach outperforms the baselines in terms of diversity and correctness. 
\end{itemize}

\begin{figure*}[t]
\centering
\includegraphics[width=\linewidth]{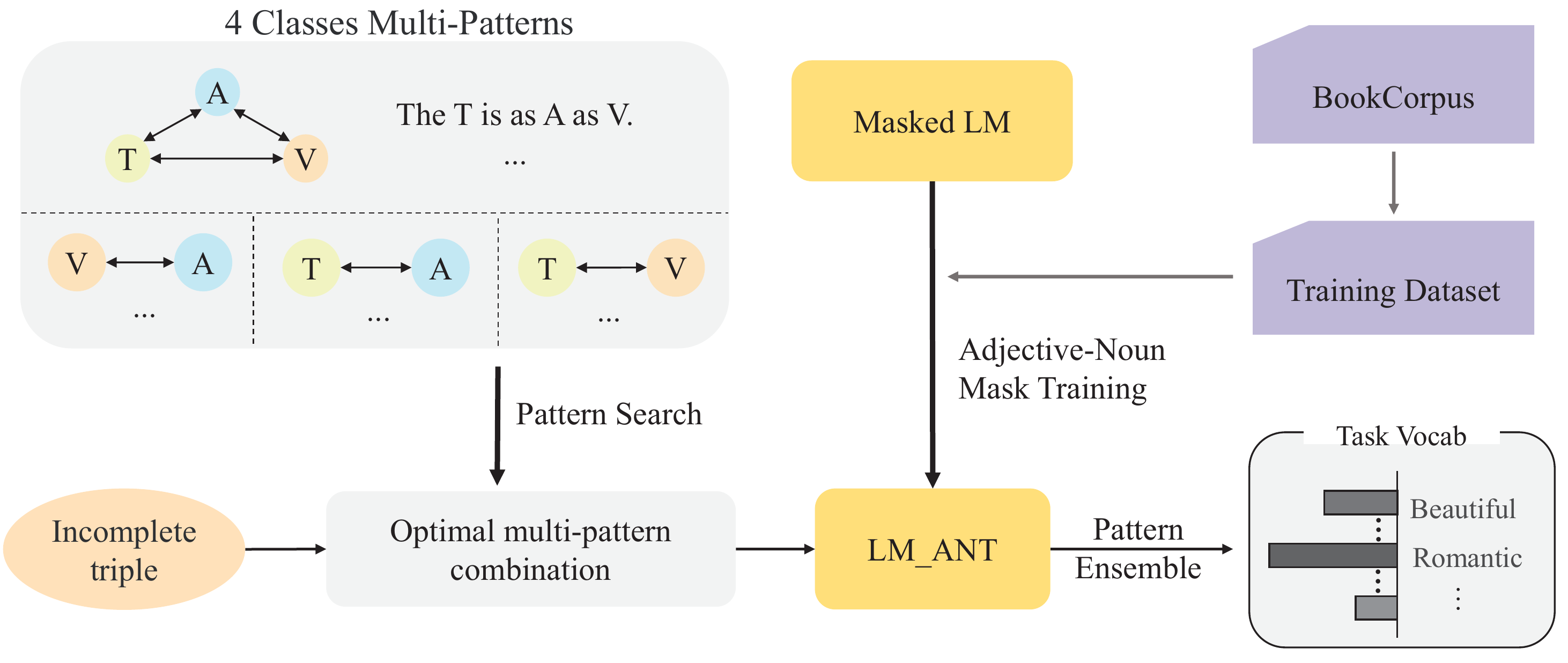}
\caption{\label{fig:framework}
The unified framework for STC. The incomplete triple will be transfer to masked sentences by a multi-pattern combination, which is searched from four classes well-designed patterns. And LM\_ANT (obtained by using unlabeled corpus to perform Adjective-Noun Mask Training) predicts the missing simile element over the task-specific vocabulary based on the masked sentences.
}
\end{figure*}

\section{Related Work}

\subsection{Simile Interpretation and Generation}
Simile interpretation and simile generation are the two main directions of the simile study \cite{Yu2019HowTA}. 
The SI task \cite{Shutova2010AutomaticMI, SuHC17} aims at finding a suitable attribute when given the tenor and vehicle, while the SG task \cite{Yu2019HowTA} is to find a proper vehicle when given the tenor and its attribute. For simile interpretation,  some works \cite{xiaobing, bar2018metaphor, meta4meaning, aglianoPBH16, QadirRW16} applied word vectors to decide which attribute words can fit into the tenor and vehicle domains and some other works \cite{Metaphoria, StoweCPMG20} introduced knowledge base \cite{framenet, conceptnet} to help find intermediate attributes. For simile generation, some works focused on constructing limited training corpus to finetune a sequence-to-sequence model \citep{Bart}  by pattern-based \citep{xiaomi, bollegala2013metaphor} or knowledge-based approaches \citep{ChakrabartyMP20, ChakrabartyZMP21, StoweCPMG20}. There are also some works \citep{Abe_acomputational, HervasCCGP07, xiaobing} that focused more on the relationships between concepts (i.e., tenor and vehicle) and attribute.
However, our paper carries out the task of simile interpretation and generation uniformly in the form of simile triples. And instead of extracting the simile triples from the limited corpus using designed templates or a hand-crafted knowledge base, we probe simile-related knowledge from PLMs.

\subsection{Explore knowledge from PLMs}
Pre-trained language models such as Bert and GPT \citep{devlin2018bert, gpt2} are trained on the large-scale unlabeled corpus. Many recent works \citep{ManningCHKL20, Ettinger20, LAMA, ShinRLWS20, HavivBG21, JiangXAN20, ZhongFC21, Wang2022ClidSumAB, Wang2022IncorporatingCK, LiL20}
focused on exploring the rich knowledge embedded in these PLMs. \citet{ManningCHKL20} and \citet{Ettinger20} learned the syntactic and semantic knowledge from PLMs. Among these works, one branch of works\citep {LAMA, ShinRLWS20,HavivBG21, JiangXAN20} designed discrete patterns to explore the common sense and world knowledge embedded in PLMs. In addition, some works \citep{ZhongFC21,LiL20} probed knowledge by searching the best-performing continuous patterns in the space of embedding. Inspired by the above works, in this paper, we probe the knowledge of simile in these pre-trained models and further apply pattern ensemble and pattern search to improve the results.

\section{Backbone}

\subsection{Simile Triple Completion} \label{std}

As shown in Figure~\ref{fig1}, the simile triple complete consists of two tasks: simile interpretation (SI) and simile generation (SG). Each simile sentence can be abstracted into the form of a triple. Therefore, we define a triple: $(\mathcal T, \mathcal A, \mathcal V)$, where $\mathcal T$, $\mathcal V$ are mainly nouns or noun phrases and represent the tenor and vehicle in the simile sentence, respectively. $\mathcal A$ is the attribute in simile sentences, which is an adjective. If the $\mathcal A$ is None in the triple, i.e. $(\mathcal T, None, \mathcal V)$, we define it as the simile interpretation task. Similarly, if the $\mathcal V$ is None, i.e. $(\mathcal T, \mathcal A, None)$, this will be the task of simile generation.

\subsection{Masked Language Model}
The masked language model (MLM) \citep{devlin2018bert, Taylor1953ClozePA} randomly masks the words in the input sentence and feeds the masked sentence into the pre-trained models to make predictions by other visible words. For example, given a sentence $s = [w_1, w_2, \dots, w_i, \dots, w_m]$, where the $w_i$ means the $i$-th word in the sentence. We can randomly mask $s$ and feed the masked sequence $\widetilde {s}$ into the PLMs e.g. BERT \citep{devlin2018bert} to obtain the masked words by Equation:

\begin{equation}
    \widetilde{s} = f_{mask}(s, i, v) 
    \label{mlm}
\end{equation}

\begin{equation}
    P = f_\theta (\widetilde{s} )
    \label{mlm1}
\end{equation}

where the $v$ means the Vocabulary for pre-trained models, and the $i$ denotes the position of the masked word in Equation~\ref{mlm}. The $\theta$ is the parameters of PLMs in Equation~\ref{mlm1}. We can select the word corresponding to the maximum probability in $P$ as the output of the model.

\subsection{Probe Simile Knowledge with MLM} \label{prob}
To probe the simile knowledge in pre-trained masked language models, the intuitive solution is: (1) Construct a sentence that contains the simile triple in Section~\ref{std} with the given pattern. (2) Mask the attribute $\mathcal A$ or vehicle $\mathcal V$ in this simile sentence. (3) Predict the words in the masked position with MLM. For example, when given a pattern \textit{The $\mathcal T$ is as $\mathcal A$ as $\mathcal V$}, the input sentence of MLM is \textit{The $\mathcal T$ is as [MASK] as $\mathcal V$} for the SI task while \textit{The $\mathcal T$ is as $\mathcal V$ as [MASK]} for the SG task.

To formulate this problem, we define the pattern function as $p(\tau)$, where $\tau \in \{SG, SI\}$. The pre-trained MLM is denoted as $\mathcal M$ and the predicted distribution $Q$ over vocabulary $V$ can be formulated as:

\begin{equation}
    Q(w|p(\tau)) = \frac{exp({\mathcal M(w|p(\tau)))}}{\sum_{w' \in V}exp({\mathcal M(w'|p(\tau)))}}
    \label{softmax}
\end{equation}

\begin{table*}[t]
\small
\centering
\begin{tabular}{cclc}
\hline
Class                 & Relationship                                  & Pattern        & ~                                                     \\ \hline
\multirow{3}{*}{\rnum{1}} 
&\multirow{3}{*}{
\begin{minipage}[b]{0.3\columnwidth}
	\centering
	\raisebox{-.5\height}{\includegraphics[width=0.8\linewidth]{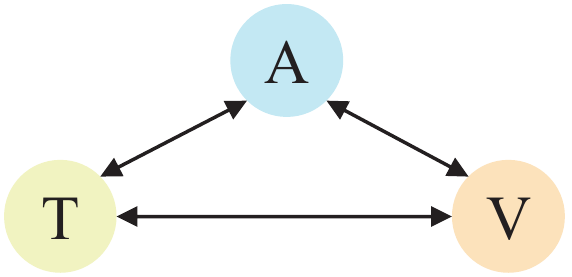}}
\end{minipage}
}

 & The \{tenor\} is as \{attribute\} as \{vehicle\}.   & $p_1$                 \\
                      &                                            & \{vehicle\} is very \{attribute\}, so as \{tenor\}.    & $p_2$              \\
                      &                                             & \{tenor\} is like \{vehicle\}, because they are both \{attribute\}. & $p_3$\\ \hline
\multirow{3}{*}{\rnum{2}} 
& \multirow{3}{*}{
\begin{minipage}[b]{0.2\columnwidth}
	\centering
	\raisebox{-.5\height}{\includegraphics[width=\linewidth]{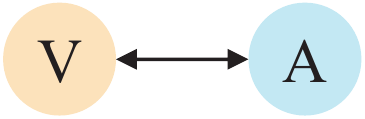}}
\end{minipage}
}

 & The \{attribute\} \{vehicle\}. & $p_4$                                     \\
                      &                                            & \{vehicle\} is very \{attribute\}.   & $p_5$                                \\
                      &                                            & \{vehicle\} is \{attribute\}.   & $p_6$                                     \\ \hline 
\multirow{3}{*}{\rnum{3}} 
& \multirow{3}{*}{
\begin{minipage}[b]{0.2\columnwidth}
	\centering
	\raisebox{-.5\height}{\includegraphics[width=\linewidth]{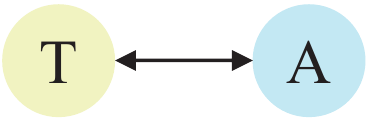}}
\end{minipage}
}
 & The \{attribute\} \{tenor\}.  & $p_7$                                      \\
                      &                                            & \{tenor\} is very \{attribute\}.   & $p_8$                                  \\
                      &                                             & \{tenor\} is \{attribute\}.                   & $p_9$                      \\ \hline
\multirow{3}{*}{\rnum{4}} 
& \multirow{3}{*}{
\begin{minipage}[b]{0.2\columnwidth}
	\centering
	\raisebox{-.5\height}{\includegraphics[width=\linewidth]{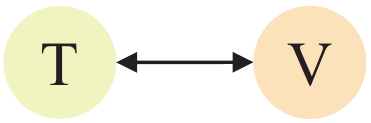}}
\end{minipage}
}
& \{tenor\} is similar to \{vehicle\}.  & $p_{10}$                               \\
                      &                                            & \{tenor\} is like \{vehicle\}.        & $p_{11}$                               \\
                      &                                            & \{tenor\} and \{vehicle\} are alike.        & $p_{12}$                         \\ \hline
\end{tabular}
\caption{All patterns and corresponding classes. Class \rnum{1} models the relationship between three elements, and other classes model relationships between two elements. Every pattern is denoted as the right side symbol $p_i$.}
\label{tab:pattern}
\end{table*}

\section{Method}
In this section, we will introduce our proposed method of probing simile knowledge from pre-trained models. Our method first introduces a secondary pre-training stage - Adjective-Noun mask Training (ANT) based on pre-trained language models to acquire diverse lexical-specific words. Then two modules of pattern ensemble and pattern search are used to obtain the high-quality predictions. The framework of our method is shown in Figure~\ref{fig:framework} in detail\footnote{we released our code at https://github.com/nairoj/Probing-Simile-from-PLM.}.

\subsection{Adjective-Noun Mask Training (ANT)} 

For the MLM task, pre-trained models prefer to output high-frequency words as candidate words since the objective of the training is to minimize the cross-entropy loss \citep{Gehrmann2019GLTRSD}. However, the components of simile triples are usually nouns or adjectives and the simile sentences are appealing due to their creativity and unexpectedness. Therefore, to predict more diverse and specific words of simile component, we introduce a secondary pre-training stage - Adjective-Noun mask Training (ANT) that fine-tune the pre-trained model with specially designed datasets. First, we utilize \textit{trankit} \citep{van2021trankit} to construct the training set by selecting sentences from BookCorpus \citep{Zhu_2015_ICCV} that contains \textit{amod}\footnote{
An adjectival modifier of a noun (or pronoun) is any adjectival phrase that serves to modify the noun (or pronoun). The relation applies whether the meaning of the noun is modified in a compositional way (e.g., large house) or an idiomatic way (hot dogs).} dependencies \citep{Uni_Depend}. Second, we mask a word at the end of \textit{amod} relation, instead of randomly masking, and all words are masked no more than $5$ times. Finally, the pre-trained model is finetuned on the constructed dataset with MLM loss. In this way, the pre-trained model will avoid the bias to high-frequency words and have a higher probability of generating diverse and novel words. 

\subsection{Pattern Ensemble (PE)} \label{PE}
Since words predicted by MLM have a preference for different patterns and only using one pattern is insufficient, we apply the pattern ensemble to obtain better performance where different types of patterns are designed as shown in Table \ref{tab:pattern}. Specifically, the class \rnum{1} describes the relationship between the three-element $\mathcal{T}$, $\mathcal{V}$ and $\mathcal{A}$. However, the similes tend to highlight an obvious attribute between tenor and vehicle \citep{israel2004simile}. We further design the class \rnum{2} and class \rnum{3} to find the attribute corresponding to the tenor and vehicle, respectively. Finally, the attributes of simile sentences are sometimes omitted and thus the class \rnum{4} is designed to deal with this case. Additionally, we also design three patterns for each class to obtain high-quality and robust results.

The output distribution $Q_{PE}$ of pattern ensemble can be formulated as

\begin{equation}
    \label{equ:pe}
    Q_{PE}(w|P) = \frac{1}{|P|}\sum_{p(\tau) \in P}log(Q(w|p(\tau)))
\end{equation}

where $P$ is the set of patterns $p(\tau)$ for specific task $\tau$.
Note that though we design four classes of patterns in Table \ref{tab:pattern}, some classes of patterns are not required for the SI or SG task. Specifically, The patterns of Class \rnum{4} are not used for the SI task because the attribute $\mathcal A$ is missed in Class \rnum{4}. Likewise, the patterns of Class \rnum{3} are not used for the SG task due to the lack of vehicle $\mathcal V$.

\subsection{Pattern Search (PS)} \label{PS}
The prediction of pattern ensemble in Section \ref{PE} is averaged by adding up the output distributions of all the patterns. Conversely, the hand-designed patterns are heuristic, which may lead to suboptimal results. Therefore, it is worth studying how these patterns can be combined to obtain better performance. To solve this problem, we introduce an approach of pattern search (PS) to find the best combination of different patterns. Specifically, given a simile dataset ${\mathcal D}_{PS}$, we calculate Equation \ref{equ:pe} on ${\mathcal D}_{PS}$ by iterating all subsets of the patterns. Finally, we select the optimal subset $p_{best}$ as the input of MLM to predict simile components.

\section{Experiments}\label{experiment}

\subsection{Dataset} \label{dataset}
\textbf{Dataset for ANT:} 
We constructed our train set of ANT from BookCorpus. We first extracted the sentences with length less than 64 and then masked nouns or adjectives in them based on \textit{amod} dependencies \citep{Uni_Depend}. Meanwhile, we limited the frequency of masked words to less than 5. Finally, we got 98k sentences as the dataset of ANT, which contains 68k noun-masked sentences and 30k adjective-masked sentences.

\textbf{Dataset for PE and PS:}
We evaluate our method on the dataset proposed in \citep{roncero2015semantic}. As the samples in Table~\ref{tab:sample}, there are multiple attributes for each ($\mathcal T$, $\mathcal V$) pair. For example, the pair of (anger, fire) has the attributes of dangerous, hot, and red. In addition, we followed the previous work \citep{meta4meaning} to filter the dataset by reversing simile triples with attribute frequencies greater than 4. Eventually, we obtain the train set with 533 samples and the test set with 145 samples. Notice that the train set is the ${\mathcal D}_{PS}$ in Section~\ref{PS} used for the pattern search and the test set is used for evaluating all the approaches in this paper.

\begin{table}[t]
\centering
\begin{tabular}{lc}
\hline
\multicolumn{1}{c}{Triple}                        & Frequency \\ \hline
(Anger, Dangerous, Fire)   & 8        \\
(Anger, Hot, Fire)         & 8        \\
(Anger, Red, Fire)         & 5        \\
(Love, Beautiful, Rainbow)       & 10        \\
(Love, Beautiful, Melody)    & 2        \\
(Love, Beautiful, Rose)     & 9        \\ \hline
\end{tabular}
\caption{\label{tab:sample}
Some samples from the dataset. Frequency represents the number of annotators who consider the attribute is suitable for the Tenor-Vehicle pair.}
\end{table}

\subsection{Implementation Details}
\textbf{Details for ANT}: 
In adjective-nouns mask training, we utilized Adam as our optimizer and the learning rate is 5e-5. The batch size is set to 32 and the max sequence length is set to 64, respectively. Further, we utilize the Bert-Large\footnote{https://huggingface.co/Bert-large-uncased} with 340M parameters as the basic model to perform adjective-nouns mask training and the number of training epoch is $3$.

\textbf{Vehicle Filtering}: 
For simile generation, we filter the predicted vehicles that are similar to the tensor by calculating the semantic similarity with Glove embedding. For instance, given the sentence ``The child is as tall as [MASK]", we will filter out the word ``father" as its vehicle due to not meeting the simile definition\footnote{Using something with similar characteristics to give an analogy to another thing}. To solve this problem, we compute the similarity score of the tenor and vehicle and filter the predicted vehicle whose score is above the threshold 0.48\footnote{The threshold is the maximum similarity score of tenor and vehicle in the train set}.

\begin{figure*}[t]
\centering
\includegraphics[width=\linewidth,scale=1.00]{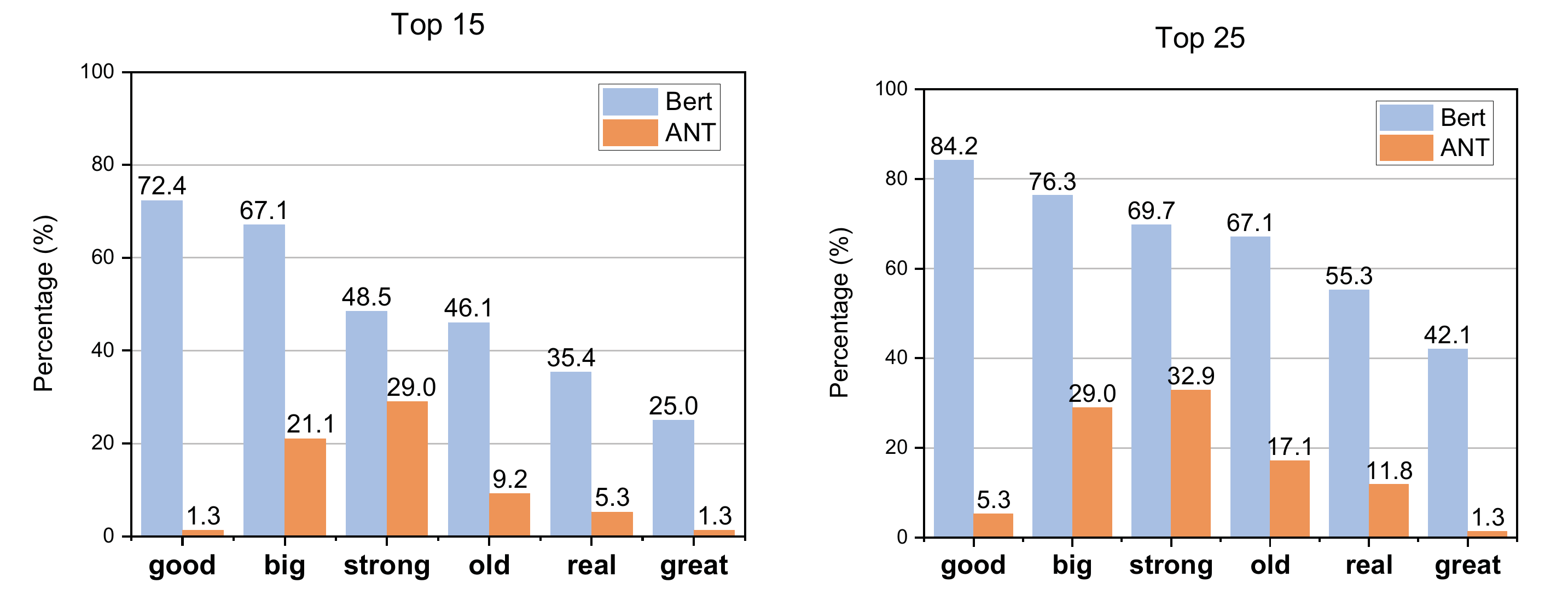}
\caption{\label{fig:diversity}
Percentage of samples whose top $K$ predicted words contain a given common word. The horizontal coordinates are some common adjectives.
}
\end{figure*}

\subsection{Evaluating the effectiveness of ANT} \label{eval_ant}

In this section, we will demonstrate that ANT could improve the diversity of predicted words for both the SI and SG tasks. We compare the predicted results of MLM (i.e., Bert) before and after ANT, which use the patterns ``The $\mathcal T$ is as [MASK] as $\mathcal V$" for the SI task and ``The $\mathcal T$ is as $\mathcal A$ as [MASK]" for the SG task.

\textbf{Metric:} 
We evaluate the diversity of the MLM predictions by calculating the proportion of unique words in the predicted Top $K$ results on the test set. It can be formulated as

\begin{equation}
    p@K = \frac{Num(Unique\_words)}{K * N}
    \label{pk}
\end{equation}

where the $Num(Unique\_words)$ means the number of unique words, and the N represents size of the test set.

\textbf{Result:} To illustrate the effectiveness of ANT, We evaluate the results on the test set based on Equation~\ref{pk}. As shown in Table~\ref{tab:ant_res}, the diversity of predicted words significantly improves after ANT for different $p@K$, specifically about 100\% improvement for the SI task and about 50\% for the SG task. Additionally, Figure \ref{fig:diversity} plots the percentage of samples on the test set, where a given common word (e.g., good, big, strong) appears in the list of the top $k=15,25$ predicted words. We can observe that the frequency of common words decreases significantly after ANT. For example, the frequency of the common word \textit{good} decreases from $72.37\%$ to $1.32\%$ when $k=15$.

\begin{table}[ht]
\centering

\resizebox{\columnwidth}{!}{
\begin{tabular}{cccccc}
\hline
  & Method & p@5 & p@10 & p@15 & p@25 \\ \hline
 & Bert & 0.263 & 0.216 & 0.189 & 0.163 \\
\multirow{-2}{*}{SI} & ANT & 0.492 & 0.412 & 0.382 & 0.312 \\ \hline
 & Bert & 0.232 & 0.201 & 0.182 & 0.158 \\
\multirow{-2}{*}{SG} & ANT & 0.370 & 0.299 & 0.256 & 0.216 \\ \hline
\end{tabular}}
\caption{\label{tab:ant_res}
The results of diversity on both the SI and SG tasks. The method \textit{Bert} and \textit{ANT} separately represent the results before and after the Adjective-Noun mask training.
}
\end{table}

\subsection{Evaluating the effectiveness of PE and PS}

\begin{table*}[ht]
\centering
\begin{tabular}{clcccccc}
\hline
\multicolumn{1}{l}{Task} & Method       & MRR            & {\color[HTML]{000000}  R@5} & {\color[HTML]{000000}  R@10} & {\color[HTML]{000000}  R@15} & {\color[HTML]{000000}  R@25} & {\color[HTML]{000000}  R@50} \\ \hline
                         & Meta4meaning & N/A            & 0.221                           & 0.303                            & 0.339                            & 0.397                            & 0.454                            \\
                         & GEM          & \textbf{0.312}          & 0.198                           & 0.254                            & 0.278                            & 0.405                            & 0.562                            \\
                         & ConScore     & 0.078          & 0.076                          & 0.138                            & 0.172                            & 0.269                            & 0.386                            \\ 
                         & Bert         & 0.266          & \underline{0.338}& \underline{0.428}                            & 0.448                            & 0.538                            & \underline{0.641}                            \\ \cline{2-8}
                         & ANT          & 0.245 & 0.310                   & 0.407                   & \underline{0.455}                            & 0.510                             & 0.614                            \\
                         & ANT+PE       & 0.241          & 0.331                           & 0.400                              & 0.448                            & \underline{0.552}                  & 0.628                            \\
\multirow{-7}{*}{SI}     & ANT+PS+PE       & \underline{0.270}           & \textbf{0.379}                           & \textbf{0.490}                             & \textbf{0.524}                            & \textbf{0.579}                            & \textbf{0.655}                            \\ \hline 
                         & ConScore     & 0.036          & 0.055                           & 0.09                             & 0.103                            & 0.145                            & 0.200                              \\  
                         & Bert         & \underline{0.064}          & \underline{0.076}                           & \underline{0.124}                            & \textbf{0.159}                            & \underline{0.207}                            & 0.283                            \\ \cline{2-8}
                         & ANT          & 0.049          & 0.069                           & 0.117                            & \underline{0.145}                            & 0.186                            & \textbf{0.303}                            \\
                         & ANT+PE       & 0.036          & 0.034                           & 0.083                            & 0.097                            & 0.131                            & 0.172                            \\
\multirow{-5}{*}{SG}     & ANT+PS+PE       & \textbf{0.095}          & \textbf{0.124}                           & \textbf{0.145 }                           & \textbf{0.159}                            & \textbf{0.214}                            & \underline{0.290}                             \\ \hline
\end{tabular}
\caption{\label{tab:automatic}
Automatic evaluation for SI and SG tasks. The best results are in bold, and the second best results are underlined. }
\end{table*}

\subsubsection{Baselines}
We compare the proposed approaches with the following baseline:

(1) \textbf{Meta4meaning \citep{meta4meaning}}: It uses the trained LSA vector representation according to the degree of abstraction and salience imbalance to select appropriate attributes. 
(2) \textbf{GEM \citep{bar2018metaphor}}: A method calculates the cosine similarity and normalized PMI between each attribute and tensor/vehicle based on Glove representing to obtain the best attribute with ranking.
(3) \textbf{Bert \citep{devlin2018bert}}: Directly use pre-trained MLM to predict the simile component with a single pattern as Section \ref{prob}. In this paper, we utilize the \textit{bert-large-uncased} as the basic pre-trained MLM.
(4) \textbf{ConScore \citep{xiaobing}}: A connecting score is proposed to select an attribute word $\mathcal A$ for $\mathcal T$ and $\mathcal V$.

Our proposed approaches are denoted as:

(1) \textbf{ANT}: Perform Adjective-Noun mask Training based on a pre-trained MLM with the datasets mentioned in Section~\ref{dataset}.
(2) \textbf{ANT+PE}: Based on ANT, the output distribution over vocabulary is predicted by average on all the corresponding patterns in Table~\ref{tab:pattern}. 
(3) \textbf{ANT+PS+PE}: Based on ANT, first the pattern search is to decide which patterns in Table~\ref{tab:pattern} are applied, and then the pattern ensemble is used over these selected patterns.


\begin{table}[t]
\resizebox{\columnwidth}{!}{
\begin{tabular}{cllll}
\hline
\multicolumn{1}{l}{Task} & \multicolumn{1}{c}{Method} & \multicolumn{1}{c}{Top5} & \multicolumn{1}{c}{Top10} & \multicolumn{1}{c}{Top15} \\ \hline
\multirow{5}{*}{SI} & ConScore & 0.192$^\dagger$ & 0.169$^\dagger$ & 0.172$^\dagger$ \\
 & Bert & 0.411$^\dagger$ & 0.364$^\dagger$ & 0.326$^\dagger$ \\ \cline{2-5} 
 & ANT & 0.471 & 0.396$^\dagger$ & 0.365$^\dagger$ \\
 & ANT+PE & \underline{0.494} & \textbf{0.469} & \textbf{0.456} \\
 & ANT+PS+PE & \textbf{0.496} & \underline{0.433}$^\dagger$ & \underline{0.398}$^\dagger$ \\ \hline 
\multirow{5}{*}{SG} & ConScore & 0.780$^\dagger$ & 0.690$^\dagger$ & 0.673$^\dagger$ \\
 & Bert & 0.597$^\dagger$ & 0.667$^\dagger$ & 0.629$^\dagger$ \\ \cline{2-5} 
 & ANT & 0.867$^\dagger$ & \underline{0.868}$^\dagger$ & \underline{0.808}$^\dagger$ \\
 & ANT+PE & \underline{0.887}$^\dagger$ & 0.805$^\dagger$ & 0.751$^\dagger$ \\
 & ANT+PS+PE & \textbf{1.123} & \textbf{1.052} & \textbf{0.973} \\ \hline
\end{tabular}}

\caption{\label{tab:human}
The average score of human evaluation for STC. The best results are in bold, and the second best results are underlined. $\dagger$ denotes significant difference with the best result (t-test, p-value<0.05).
}
\end{table}

\subsubsection{Metrics}
We use both automatic evaluation and human evaluation to compare our approaches with baselines.

\textbf{Automatic Evaluation:}

(1) Mean Reciprocal Rank (MRR): average on the reciprocal of the ranking $r_i$ of label words in the predicted candidates, denoted as
\begin{equation}
    MRR = \frac{1}{N}\sum_{i=1}^N{\frac{1}{r_i}}
    \label{mrr}
\end{equation}

(2) $R@K$: the percentage of the label words appear in the top $K$ predictions. Note that, following previous works \citep{meta4meaning, bar2018metaphor}, we consider a predicted word as the correct answer if it is a synonym of label word n in WordNet \citep{miller1995wordnet}. It can be formulated as
\begin{equation}
  cor(w)=
   \begin{cases}
     1 & {w \in Synonyms(L_i)} \\
     0 & {w \notin Synonyms(L_i)}
   \end{cases}
  \label{cov}
\end{equation}


\begin{equation}
    R@K = \frac{1}{N}\sum_{i=1}^N{\frac{\sum_{w \in K_{i}}{cor(w)}}{K}}
    \label{recal}
\end{equation}

where $K_i$ denotes the list of predicted words, $L_i$ denotes the list of label words and $Synonyms(L_i)$ represents the synonyms of a word.

\textbf{Human Evaluation: } To further prove our approaches are better than baselines, human evaluation is used to evaluate the quality of predicted simile triples from three levels (0, 1, 2). 0 - The triple is unacceptable. 1 - The triple is acceptable. 2 - The triple is acceptable and creative. Given a simile triple, annotators need to score it according to their subjective judgment and each triple is annotated by three annotators independently. We use the average score of three annotators as the quality of a simile triple.

\subsubsection{Results}

\begin{table*}[htbp]
\centering
\begin{tabular}{clccccc}
\hline
\multicolumn{1}{l}{Task} & \multicolumn{1}{c}{Subset of Patterns} & MRR &  R@5 &  R@10 &  R@15 &  R@25 \\ \hline
\multirow{4}{*}{SI} & $\{p_1, p_5\}$ & \textbf{0.100} & \textbf{0.126} & \textbf{0.184} & \textbf{0.233} & \textbf{0.281} \\
 & $\{p_1, p_2, p_3, p_4, p_5,   p_9\}$ & 0.095 & 0.107 & 0.171 & 0.203 & 0.268 \\
 & $\{p_1, p_2, p_3, p_4, p_5,   p_7, p_8\}$ & 0.095 & 0.099 & 0.163 & 0.206 & 0.274 \\
 & $\{p_1, p_4, p_5\}$ & 0.094 & 0.094 & 0.163 & 0.203 & 0.261 \\ \hline
\multirow{4}{*}{SG} & $\{p_3, p_4\}$ & \textbf{0.056} & 0.068 & \textbf{0.105} & \textbf{0.135} & \textbf{0.159} \\
 & $\{p_1, p_4\}$ & 0.056 & \textbf{0.071} & 0.092 & 0.120 & 0.154 \\
 & $\{p_1, p_3, p_4\}$ & 0.052 & 0.06 & 0.105 & 0.128 & 0.163 \\
 & $\{p_1, p_2, p_4\}$ & 0.052 & 0.058 & 0.096 & 0.116 & 0.137 \\ \hline
\end{tabular}
\caption{\label{tab:subsets}
The top 4 best performing pattern subsets for SI and SG tasks (see Table \ref{tab:pattern} for which class the pattern $p_i$ belongs to). The best results are in bold. More results of pattern search are shown in the Appendix~\ref{sec:appendix_1}.}
\end{table*}

\textbf{Automatic and Human Evaluation:}
The results of both automatic and human evaluation are shown in Table~\ref{tab:automatic} and Table~\ref{tab:human}. The agreement between annotators is measured using Fleiss’s kappa $\kappa$ \citep{randolph2005free}. The $\kappa$ value is 0.68 (substantial agreement) for the SI task and 0.48 (moderate agreement) for the SG task. 

From the results, we can conclude
\begin{itemize}
\item[(1)] For both SI and SG tasks, our proposed approaches (i.e., ANT, ANT+PE, ANT+PS+PE) significantly outperform the baselines on both automatic and human evaluations. It proves that our methods not only enhance the diversity of predicted simile components in Section \ref{eval_ant} but also their quality.

\item[(2)] Pre-trained MLM-based methods (i.e., Bert, ANT, ANT+PE and ANT+PS+PE) perform better than the traditional methods (i.e., GEM, Meta4meaning, ConScore). It shows the potential of pre-trained models in probing simile knowledge.

\item[(3)] Compared ANT with Bert, we found that though ANT improves the diversity of predicted words in Table \ref{tab:ant_res}, the average scores on automatic and human evaluations decrease because the simile knowledge is not involved in the ANT training process. However, our proposed PE and PS compensate for the performance. 

\item[(4)] The scores of automatic evaluation metrics on the SI task are remarkably higher than the SG task. Yet, the scores of human evaluation metrics are significantly lower than on the SG task. 
We conjecture that this may be because the list of candidate words of attribute predicted by SI are smaller than that of the vehicle for the SG task. 
For example, given the SI sample ``(Cloud, $None$, Cotton)'', the attribute words are almost restricted to the physical properties of the vehicle, such as ``Soft'', while the choices of vehicle words are more varied and unexpected given the SG sample ``(Cloud, soft, $None$)'' such as ``cotton, silk, towel".
\end{itemize}

\textbf{Discussion for PS:} Compared ANT+PS+PE to ANT+PE, it can be included that pattern search brings a great improvement to the results on both automatic and human evaluations. To have a deeper insight into PS, the pattern subsets with high performance are listed in Table~\ref{tab:subsets}. For the SI task, the optimal multi-pattern combination is $\{p_1, p_5\}$, which support the hypothesis proposed by \citep{ortony1979beyond} considers that the highlighted attribute of a simile triple is more salient in the vehicle domain despite it is commonly shared by both tenor and vehicle domains. Specifically, pattern $p_1$ belonging to the Class \rnum{1}, models the relationship of all three simile components while the pattern $p_5$ belonging to Class \rnum{2} requires the candidate words to be the salient attribute of the vehicle. Similarly, for SG task, optimal multi-pattern combination is $\{p_3, p_4\}$, which is also a combination of the Class \rnum{1} pattern and the Class \rnum{2} pattern. 

\section{Conclusion and Future work}
In this paper, from the perspective of simile triple completion, we propose an unified framework to solve the SI and SG tasks by probing the knowledge of the pre-trained masked language model. The backbone of our method is to construct masked sentences with manual patterns from an incomplete simile triple, and then use language models with MLM heads to predict the masked words. Moreover, a secondary pre-training stage (the adjective-noun mask training) is applied to improve the diversity of predicted words. Pattern ensemble (PE) and pattern search (PS) are further used to improve the quality of predicted words. Finally, automatic and human evaluations demonstrate the effectiveness of our framework in both SI and SG tasks. 
In future work, we will continue to study how to mine broader or complex knowledge from pre-trained models, such as metaphor, common sense and we expect more researchers to perform related research.


\section*{Acknowledgements}
This work is supported by the Key Research and Development Program of Zhejiang Province (No. 2022C01011) and National Natural Science Foundation of China (Project 61075058). We would like to thank the anonymous reviewers for their excellent feedback.


\bibliography{output}
\bibliographystyle{acl_natbib}

\appendix 

\section{More results of Pattern Search } \label{sec:appendix_1}
The more results of Pattern Search are shown in Table~\ref{tab:PS_result}.
\begin{table*}[t]
\centering
\begin{tabular}{clcccccc}
\hline
Task & \multicolumn{1}{c}{Subset of Patterns} & MRR & {\color[HTML]{000000} R@5} & R@10 & R@15 & R@25 & R@50 \\ \hline
 & $\{p_1,   p_5\}$ & 0.100 & 0.126 & 0.184 & 0.233 & 0.281 & 0.375 \\
 & $\{p_1, p_2, p_3, p_4, p_5,   p_9\}$ & 0.095 & 0.107 & 0.171 & 0.203 & 0.268 & 0.377 \\
 & $\{p_1, p_2, p_3, p_4, p_5,   p_7, p_8\}$ & 0.095 & 0.099 & 0.163 & 0.206 & 0.274 & 0.373 \\
 & $\{p_1, p_4, p_5\}$ & 0.094 & 0.094 & 0.163 & 0.203 & 0.261 & 0.366 \\
 & $\{p_1, p_2, p_4, p_5, p_9\}$ & 0.094 & 0.111 & 0.165 & 0.214 & 0.265 & 0.373 \\
 & $\{p_1, p_4, p_5, p_6, p_8\}$ & 0.093 & 0.109 & 0.171 & 0.212 & 0.283 & 0.368 \\
 & $\{p_1, p_4, p_5, p_6, p_7,   p_8, p_9\}$ & 0.093 & 0.090 & 0.152 & 0.205 & 0.263 & 0.338 \\
 & $\{p_1, p_2, p_4, p_5\}$ & 0.093 & 0.113 & 0.167 & 0.210 & 0.280 & 0.371 \\
 & $\{p_1, p_2, p_4, p_6, p_8\}$ & 0.093 & 0.111 & 0.178 & 0.218 & 0.272 & 0.371 \\
 & $\{p_1, p_2, p_4, p_5, p_6,   p_8\}$ & 0.093 & 0.105 & 0.173 & 0.216 & 0.283 & 0.370 \\
 & $\{p_1, p_2, p_4, p_5, p_6,   p_7, p_8, p_9\}$ & 0.093 & 0.096 & 0.156 & 0.210 & 0.261 & 0.347 \\
 & $\{p_1, p_3, p_4, p_5, p_8,   p_9\}$ & 0.093 & 0.098 & 0.159 & 0.223 & 0.265 & 0.368 \\
 & $\{p_1, p_5, p_6, p_7\}$ & 0.092 & 0.101 & 0.163 & 0.203 & 0.274 & 0.366 \\
 & $\{p_1, p_2, p_5, p_9\}$ & 0.092 & 0.099 & 0.171 & 0.225 & 0.285 & 0.362 \\
 & $\{p_1, p_2, p_4, p_5, p_8\}$ & 0.092 & 0.105 & 0.173 & 0.218 & 0.280 & 0.360 \\
 & $\{p_1, p_2, p_4, p_5, p_8,   p_9\}$ & 0.092 & 0.099 & 0.158 & 0.216 & 0.274 & 0.360 \\
 & $\{p_1, p_2, p_4, p_5, p_6,   p_8, p_9\}$ & 0.092 & 0.094 & 0.159 & 0.210 & 0.270 & 0.355 \\
 & $\{p_1, p_3, p_5\}$ & 0.092 & 0.105 & 0.169 & 0.216 & 0.280 & 0.381 \\
 & $\{p_1, p_3, p_4, p_5, p_8\}$ & 0.092 & 0.096 & 0.173 & 0.220 & 0.276 & 0.368 \\
 & $\{p_1, p_2, p_3, p_5, p_9\}$ & 0.092 & 0.107 & 0.165 & 0.208 & 0.281 & 0.371 \\
 & $\{p_1, p_5, p_6\}$ & 0.091 & 0.116 & 0.180 & 0.220 & 0.283 & 0.385 \\
 & $\{p_1, p_5, p_6, p_8\}$ & 0.091 & 0.111 & 0.174 & 0.229 & 0.278 & 0.364 \\
 & $\{p_1, p_4, p_5, p_8\}$ & 0.091 & 0.103 & 0.176 & 0.216 & 0.291 & 0.366 \\
\multirow{-25}{*}{SI} & $\{p_1, p_4, p_5, p_8, p_9\}$ & 0.091 & 0.099 & 0.165 & 0.205 & 0.270 & 0.358 \\
\hline
 & $\{p_3,   p_4\}$ & 0.056 & 0.068 & 0.105 & 0.135 & 0.159 & 0.223 \\
 & $\{p_1, p_4\}$ & 0.056 & 0.071 & 0.092 & 0.120 & 0.154 & 0.225 \\
 & $\{p_1, p_3, p_4\}$ & 0.052 & 0.060 & 0.105 & 0.128 & 0.163 & 0.218 \\
 & $\{p_1, p_2, p_4\}$ & 0.052 & 0.058 & 0.096 & 0.116 & 0.137 & 0.197 \\
 & $\{p_1, p_4, p_5\}$ & 0.052 & 0.064 & 0.094 & 0.114 & 0.137 & 0.203 \\
 & $\{p_3, p_4, p_{11}\}$ & 0.050 & 0.058 & 0.079 & 0.099 & 0.141 & 0.186 \\
 & $\{p_1, p_4, p_6\}$ & 0.049 & 0.051 & 0.086 & 0.105 & 0.131 & 0.197 \\
 & $\{p_3, p_4, p_5\}$ & 0.048 & 0.058 & 0.096 & 0.114 & 0.144 & 0.208 \\
 & $\{p_3, p_4, p_6\}$ & 0.048 & 0.051 & 0.094 & 0.109 & 0.135 & 0.199 \\
 & $\{p_1, p_3, p_4, p_5\}$ & 0.048 & 0.049 & 0.092 & 0.120 & 0.148 & 0.208 \\
 & $\{p_1, p_3, p_4, p_6\}$ & 0.048 & 0.054 & 0.090 & 0.111 & 0.137 & 0.214 \\
 & $\{p_1, p_3, p_4, p_{11}\}$ & 0.048 & 0.062 & 0.088 & 0.105 & 0.128 & 0.188 \\
 & $\{p_2, p_3, p_4\}$ & 0.047 & 0.062 & 0.090 & 0.105 & 0.133 & 0.197 \\
 & $\{p_1, p_2, p_4, p_6\}$ & 0.047 & 0.051 & 0.084 & 0.113 & 0.146 & 0.184 \\
 & $\{p_1, p_2, p_4, p_5\}$ & 0.047 & 0.054 & 0.083 & 0.113 & 0.141 & 0.188 \\
 & $\{p_1, p_2, p_3, p_4\}$ & 0.046 & 0.058 & 0.088 & 0.109 & 0.133 & 0.206 \\
 & $\{p_1, p_2, p_3, p_4, p_5\}$ & 0.046 & 0.054 & 0.083 & 0.096 & 0.131 & 0.188 \\
 & $\{p_4, p_{11}\}$ & 0.046 & 0.053 & 0.081 & 0.099 & 0.122 & 0.171 \\
 & $\{p_1, p_3, p_4, p_5, p_{12}\}$ & 0.046 & 0.058 & 0.079 & 0.094 & 0.114 & 0.171 \\
 & $\{p_1, p_4, p_{11}\}$ & 0.045 & 0.053 & 0.084 & 0.101 & 0.139 & 0.208 \\
 & $\{p_1, p_2, p_3, p_4, p_{11}\}$ & 0.045 & 0.060 & 0.084 & 0.099 & 0.118 & 0.169 \\
 & $\{p_1, p_3, p_4, p_5, p_6\}$ & 0.045 & 0.047 & 0.083 & 0.116 & 0.137 & 0.184 \\
 & $\{p_1, p_4, p_5, p_6\}$ & 0.045 & 0.049 & 0.079 & 0.101 & 0.133 & 0.189 \\
\multirow{-24}{*}{SG} & $\{p_1, p_4, p_5, p_{11}\}$ & 0.045 & 0.045 & 0.077 & 0.096 & 0.133 & 0.186 \\ \hline
\end{tabular}
\caption{\label{tab:PS_result}
The top 25 best performing pattern subsets for SI and SG tasks, sorted according to MRR. See Table \ref{tab:pattern} for which class the pattern $p_i$ belongs to. 
}
\end{table*}

\section{More Prediction}
Some results are shown in Table~\ref{tab:SI_sample} and Table~\ref{tab:SG_sample}.

\begin{table}[ht]
\centering
\begin{tabular}{ll}
\hline
\multicolumn{1}{c}{Triple} & \multicolumn{1}{c}{Score} \\ \hline
(anger,   burning, fire) & 2.00 \\
(cities, humid, jungles) & 2.00 \\
(clouds, fluffy, cotton) & 2.00 \\
(deserts, hot, ovens) & 2.00 \\
(exams, tough, hurdles) & 2.00 \\
(families, powerful,   fortresses) & 2.00 \\
(fingerprints, accurate,   portraits) & 2.00 \\
(highways, crooked, snakes) & 2.00 \\
(love, pure, flower) & 2.00 \\
(anger, blazing, fire) & 1.67 \\
(love, romantic, melody) & 1.67 \\
(money, valuable, oxygen) & 1.67 \\
(obligations, binding,   shackles) & 1.67 \\
(teachers, creative,   sculptors) & 1.67 \\
(time, important, money) & 1.67 \\
(tv, addicted, drug) & 1.67 \\
(wisdom, infinite, ocean) & 1.67 \\
(desks, messy, junkyards) & 1.33 \\
(eyelids, close, curtains) & 1.33 \\
(god, benevolent, parent) & 1.33 \\
(music, soothing, medicine) & 1.33 \\
(skating, relaxing, flying) & 1.33 \\
(friendship, lovely, rainbow) & 1.00 \\
(life, challenging, journey) & 1.00 \\
(love, sweet, flower) & 1.00 \\
(love, fragile, rose) & 1.00 \\
(pets, annoying, kids) & 1.00 \\
(television, attractive,   candy) & 1.00 \\
(women, quiet, cats) & 1.00 \\
(trust, secure, glue) & 0.67 \\
(tv, harmful, drug) & 0.67 \\
(tree trunks, weak, straws) & 0.67 \\
(trees, sturdy, umbrellas) & 0.67 \\
(winter, long, death) & 0.33 \\
(tongues, spicy, fire) & 0.33 \\
(typewriters, obsolete,   dinosaurs) & 0.00 \\
(time, quick, snail) & 0.00 \\
(trees, long, umbrellas) & 0.00 \\
(tv, ineffective, drug) & 0.00 \\
(tv, unreliable, drug) & 0.00 \\ \hline
\end{tabular}
\caption{\label{tab:SI_sample}
Some results of simile interpretation. Score is the average score of human evaluation. }
\end{table}

\begin{table}[ht]
\begin{tabular}{ll}
\hline
\multicolumn{1}{c}{Triple} & \multicolumn{1}{c}{Score} \\ \hline
(clouds,   white, cream) & 2.00 \\
(friendship, colorful,   jewelry) & 2.00 \\
(love, colorful, coral) & 2.00 \\
(love, shiny, pearl) & 2.00 \\
(skating, relaxing, noon) & 2.00 \\
(tv, addictive, drug) & 2.00 \\
(dreams, clear, crystal) & 1.67 \\
(friendship, colorful,   sunrise) & 1.67 \\
(love, addictive, coke) & 1.67 \\
(love, colorful, sunrise) & 1.67 \\
(music, cure, lullaby) & 1.67 \\
(clouds, white, pearl) & 1.33 \\
(dreams, clear, glass) & 1.33 \\
(exams, challenging, boxing) & 1.33 \\
(friendship, colorful,   pottery) & 1.33 \\
(knowledge, important, faith) & 1.33 \\
(love, addictive, alcohol) & 1.33 \\
(love, colorful, lavender) & 1.33 \\
(music, cure, art) & 1.33 \\
(clouds, white, dove) & 1.00 \\
(desks, messy, nightmare) & 1.00 \\
(desks, messy, storage) & 1.00 \\
(highways, long, march) & 1.00 \\
(knowledge, important, time) & 1.00 \\
(love, addictive, poison) & 1.00 \\
(love, colorful, perfume) & 1.00 \\
(love, colorful, silk) & 1.00 \\
(music, cure, time) & 1.00 \\
(skating, relaxing, outdoors) & 1.00 \\
(typewriters, ancient, legend) & 1.00 \\
(cities, crowded, blast) & 0.67 \\
(knowledge, important,   intuition) & 0.67 \\
(love, colorful, neon) & 0.67 \\
(clouds, white, bone) & 0.33 \\
(friendship, colorful,   lightning) & 0.33 \\
(love, addictive, spice) & 0.33 \\
(cities, crowded, hell) & 0.00 \\
(clouds, white, steel) & 0.00 \\
(dreams, clear, stone) & 0.00 \\
(exams, challenging, robotics) & 0.00 \\ \hline
\end{tabular}
\caption{\label{tab:SG_sample}
Some results of simile generation. Score is the average score of human evaluation. }
\end{table}

\end{document}